\documentclass{article}

    \PassOptionsToPackage{numbers, compress}{natbib}

\usepackage{graphicx}
\usepackage{subcaption}
\usepackage{microtype}
\usepackage{booktabs} 
\usepackage{diagbox}
\usepackage{hyperref}
\usepackage{amsmath}
\usepackage{amssymb}
\usepackage{mathtools}
\usepackage{amsthm}
\usepackage{listings}
\usepackage{xcolor}
\usepackage{inconsolata}
\usepackage{makecell} 

\definecolor{keycolor}{RGB}{0,0,132}     
\definecolor{stringcolor}{RGB}{140,40,40} 
\definecolor{commentcolor}{RGB}{90,90,90} 

 \usepackage[preprint]{neurips_2026}

\usepackage[utf8]{inputenc} 
\usepackage[T1]{fontenc}    
\usepackage{hyperref}       
\usepackage{url}            
\usepackage{booktabs}       
\usepackage{amsfonts}       
\usepackage{nicefrac}       
\usepackage{microtype}      
\usepackage{xcolor}         

\title{Concurrency without Model Changes:  \\
    Future-based Asynchronous Function Calling for LLMs}

\author{
  Guangyu Feng$^{1}$ \quad
  Huanzhi Mao$^{1}$ \quad
  Prabal Dutta$^{1}$ \quad
  Joseph E. Gonzalez$^{1}$ \\
  $^{1}$University of California, Berkeley
}

\begin{document}

\maketitle

\begin{abstract}
  Function calling, also known as tool use, is a core capability of modern LLM agents but is typically constrained by synchronous execution semantics. Under these semantics, LLM decoding is blocked until each function call completes, resulting in increasing end-to-end latency. In this work, we introduce AsyncFC, a pure execution-layer framework that decouples LLM decoding from function execution, enabling overlap between model decoding and function execution as well as inter-function parallelism when dependencies permit. AsyncFC layers over existing models and unmodified function implementations, requiring no fine-tuning or changes to the standard synchronous function-calling protocol. On coding tasks, AsyncFC achieves a 1.4$\times$ end-to-end speedup while preserving task accuracy. Furthermore, these results reveal that LLMs possess a native capability to reason over symbolic futures that represent unresolved execution results, enabling an asynchronous paradigm for model-tool interaction.
\end{abstract}

\section{Introduction}
\label{Intro}
Function calling (FC) has become a standard capability in modern large language model (LLM) agents \citep{Pantiukhin2025AcceleratingES, Liu2024ToolACEWT, Wang2023VoyagerAO, wang2026function}. It enables models to interact with external APIs, retrieve information, and take actions in the world \citep{an2025empowering, Erdogan2024TinyAgentFC,huang2025crmarena, 
Moon2024EfficientAS, Packer2023MemGPTTL, Patil2023GorillaLL, schick2023toolformer}. Modern LLM function calling follows a standard synchronous function calling protocol: the model emits a structured function call, the runtime executes it, and the result
is appended to the conversation history before the model resumes decoding. This process repeats until the model produces a final response, and each function call must be immediately followed by its corresponding result in the conversation history~\citep{google_gemini_api_function_calling, openai_function_calling, Ruan2023IdentifyingTR, qin2023toolllm, yao2022react}.

In practice, most systems implement this workflow with synchronous execution semantics. After emitting a function call, model decoding blocks until the function execution completes.
This serialization limits concurrency and increases end-to-end latency.
Prior approaches address this bottleneck by exposing in-progress function
executions to continued model generation or by shifting concurrency management
complexity onto the model side, often altering the FC protocol
or requiring model changes~\citep{gim2024asynchronous, kim2024llm,xu2023rewoo}.

In contrast, we argue that managing execution concurrency should be offloaded from the model to the system runtime. 
Inspired by asynchronous programming \citep{baker1977futures, liskov1988promises}, we introduce AsyncFC, an execution-layer framework with automatic schema adaptation that decouples model decoding from function execution.
This design is guided by our observation that LLMs can natively operate over symbolic future placeholders representing unresolved execution results. AsyncFC enables asynchronous function calling while preserving the standard synchronous call–return semantics the model expects, requiring no modifications to the underlying models or function implementations.

\begin{figure}[t]
  \begin{center}
  \centerline{\includegraphics[width=\columnwidth]{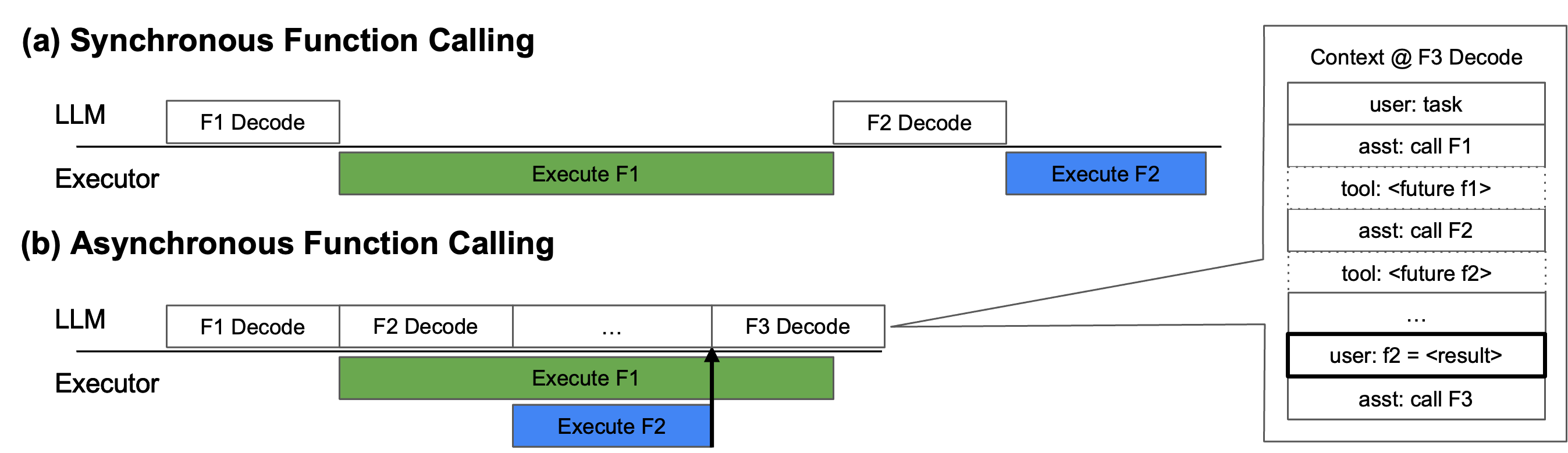}}
    \caption{
    \textbf{Timeline of synchronous and asynchronous function calling.} F1 and F2 are independent function calls, while F3 depends on the result of F2. The example is illustrated using a sequential function-calling API. 
    (a) Under synchronous function calling, decoding is blocked until each function execution completes.
    (b) AsyncFC returns a future placeholder immediately after dispatch, allowing decoding and function execution to overlap. Function results are integrated into the context at turn boundaries. Dashed boxes and solid bold boxes denote future placeholders and resolved results, respectively; $f_i$ denotes the future returned by $F_i$.
    }
  \end{center}
\vspace{-5mm}
\end{figure}
To achieve this, AsyncFC exposes futures to the model by automatically transforming synchronous function schemas to support future-valued inputs and outputs, and by introducing a lightweight \texttt{await\_future} function that allows the model to resolve futures. Upon receiving a function call, AsyncFC dispatches the underlying synchronous function implementation to a backend executor and immediately returns future placeholders to the LLM. By satisfying the expected call–return protocol, AsyncFC allows the model to continue decoding as if the function call had already returned, while the underlying function executes asynchronously in the background. Once function execution completes, results are integrated into the context at turn boundaries. Together, these capabilities accelerate the generation of subsequent function calls, thereby maximizing exploitable parallelism. To safely execute functions in parallel, AsyncFC features a scheduler that conservatively serializes execution by default, but leverages developer-supplied resource interaction annotations to unlock dependency-safe inter-function parallelism. As a result, AsyncFC enables two forms of concurrency: overlap between decoding and function execution, and dependency-safe inter-function parallelism.

Beyond accelerating the agentic workflow, AsyncFC serves as a framework to explore an LLM's reasoning capability over unresolved future representations. AsyncFC demonstrates that the model can natively operate over symbolic placeholder representations and adapt to the resolved outputs integrated in later turns. This capability—demonstrated without requiring any additional fine-tuning—brings a new interaction pattern between the model and the functions.

Because AsyncFC operates purely at the execution layer, it is applicable across off-the-shelf synchronous models. It enables models restricted to sequential function calling to achieve substantial inter-function parallelism and exceed the speedup of the parallel function-calling modes that lack such execution-layer support. 
Furthermore, when applied on top of models supporting native parallel function calling, AsyncFC delivers additional latency reductions beyond their native parallelism. It also provides explicit execution-order semantics that existing parallel function-calling APIs often leave underspecified, reducing failure modes caused by race conditions among same-turn stateful calls. Moreover, because AsyncFC represents all long-latency operations using the same future-based execution interface, it naturally extends beyond external functions to asynchronous thinking. 
Unlike asynchronous thinking approaches that train models to explicitly fork and join subqueries~\citep{chi2025era}, AsyncFC exposes this parallelism through execution-layer changes alone.

In this paper, we design, implement, and evaluate AsyncFC, a protocol-compatible execution layer that brings asynchronous programming into LLM function calling. 
AsyncFC combines future-based decoding, event-driven result integration, and dependency-aware scheduling while preserving the standard function-calling protocol. 
AsyncFC delivers consistent end-to-end speedups across tool-use, software engineering, and multi-hop reasoning benchmarks, without sacrificing accuracy.
Under realistic function execution latencies, it achieves a 1.26$\times$ speedup on adapted Berkeley Function Calling Leaderboard (BFCL) v4 Web Search workloads \citep{patil2025berkeley} and a 1.40$\times$ speedup on SWE-bench Lite~\citep{jimenez2023swe} when integrated with SWE-agent \citep{yang2024swe}. AsyncFC further extends to multi-hop reasoning workloads such as HotpotQA \citep{yang2018hotpotqa}.

\section{Related Work}
\label{sec:related_work}

\textbf{Sequential and Parallel FC.} 
Sequential FC restricts each model turn to issuing at most one function call and admits no concurrency. Parallel FC improves Sequential FC by allowing the model to issue multiple function calls, exposing within-turn parallelism but still blocking subsequent decoding until all same-turn functions return~\citep{openai_function_calling, google_gemini_api_function_calling}. However, existing Parallel FC APIs leave ambiguity in function execution order. Some model providers encourage concurrent function execution but do not specify the handling of partial failure and race conditions from
concurrent execution; others imply serial execution in model-emitted order through example code rather than an explicit semantic contract, avoiding race conditions but weakening the latency benefit of Parallel FC. AsyncFC addresses this gap by making execution semantics explicit. The runtime uses dependency annotations to determine overlap and serialization across calls. On Sequential FC models, AsyncFC recovers Parallel-FC-like concurrency by allowing independent calls across turns to execute concurrently. On Parallel FC models, AsyncFC serializes conflicting calls in model-emitted order while executing dependency-safe calls concurrently. Given the model-emitted call order, this makes state updates, function returns, and the resulting model context deterministic. Its inter-turn function parallelism can also recover parallelism missed by conservative model-side grouping. In both settings, AsyncFC additionally enables decode--execution overlap.

\textbf{Planning-based FC.} Planning-based approaches generate a plan or dependency graph before execution, then
execute independent calls in parallel~\citep{kim2024llm,xu2023rewoo}. These approaches can
expose cross-turn parallelism, but depart from the standard turn-by-turn function-calling protocol and shift dependency reasoning into the model, often requiring model modification or additional training. They also remain synchronous and block decoding until function execution completes. By contrast, AsyncFC achieves cross-turn parallelism, preserves the standard function calling protocol, derives dependencies online, and adds a complementary source of concurrency by overlapping model decoding with in-flight function execution, all without model modification.

\textbf{Asynchronous Reasoning Approaches.}
Asynchronous reasoning approaches allow the model to continue decoding while function executions are in flight and integrate the function results after they become available \citep{gim2024asynchronous, ginart2024asynchronous}. However, they expose incomplete function executions to the model’s reasoning process, depart from the standard function-calling protocol, and require the model to reason about whether new calls are safe while previous calls remain pending. This often requires model-side protocol changes or explicit model modification. Additionally, these approaches generally do not expose function results as futures or promises, preventing unresolved results from being passed as arguments to subsequently decoded function calls. In contrast, AsyncFC shifts responsibility for enforcing dependency-safe function execution from the model to a dependency-aware scheduler. Incomplete function executions are abstracted behind symbolic futures, allowing the model to treat each function as logically complete in its context and thereby avoiding model modification. 
Furthermore, these symbolic futures can be passed as arguments during decoding, enabling additional decode–execution overlap.

\section{Method}
AsyncFC extends value-based model–function interaction to future-based interaction. To expose these futures to the model, AsyncFC automatically transforms synchronous function schemas. The runtime dispatches decoded calls asynchronously, immediately returns symbolic future identifiers to the model, and later integrates resolved results into the model context. Internally, AsyncFC tracks future-execution associations and uses a dependency-aware scheduler to execute functions concurrently when dependencies permit.

\subsection{Non-Blocking Decoding}\label{subsec:Nonblocking_Decoding}
Standard function-calling protocol requires each function invocation to be immediately followed by its return value in the conversation history, effectively blocking decoding until execution completes. 
To remove blocking, AsyncFC dispatches the synchronous function decoded by the model to an asynchronous backend executor and immediately returns to the model a future placeholder. By providing an immediate return, this placeholder satisfies protocol constraints.

To make the return value consistent with the function schema exposed to the model, AsyncFC automatically transforms function schemas to support returning a single future object as the function output. This allows the symbolic future returned by the runtime to serve as the function’s immediate return value, so the model can continue decoding subsequent calls as if the function call had already returned. To further maximize decoding before execution finishes, 
AsyncFC automatically transforms function inputs in the schema to accept either concrete values or future identifiers. This dual-typed schema allows the model to leverage its native capability to reason over future placeholders and decode subsequent calls with unresolved futures as arguments. 

In many cases, only a subset or specific fields of a function’s return value are needed as inputs to subsequent calls. To support this, AsyncFC allows developers to provide output schema annotations in example return values. When provided, the runtime automatically converts these examples into structured representations where concrete fields are replaced by symbolic future identifiers, effectively preserving the structure of the underlying function's return. Returning this structure to the model as the immediate return value gives the model fine-grained access to pending execution results when constructing downstream calls. In all transformations, function descriptions, parameter types, and return definitions are updated at the schema level, requiring no modifications to the underlying function implementation. In addition to transformed schemas, AsyncFC exposes an auxiliary function, \texttt{await\_future}. This function allows the model to explicitly request the resolution of one or more futures when concrete values are required, providing the LLM a clear reasoning trajectory to the final answer and preventing the issuance of repetitive tool calls. 

\subsection{Retrieve Function Execution Results}
\label{subsec:Retrieve_results}

While futures enable non-blocking decoding, the model must eventually retrieve the function execution results. A naive design where futures are always explicitly awaited via \texttt{await\_future} introduces additional function calls and decoding turns. To mitigate this, AsyncFC proactively integrates resolved function outputs, potentially before explicit awaits become necessary. When completed function results are available, AsyncFC integrates them into the model context at turn boundaries by inserting user messages that bind future identifiers to their corresponding resolved values.

When further model decoding strictly depends on unintegrated function execution results, the model has no option but to emit an \texttt{await\_future} call. Decoding this function incurs latency. This latency becomes overhead when the future required for subsequent decoding has already resolved after the previous turn boundary, since the runtime must wait for the call to be fully decoded before integrating the already-available result. To reduce the latency penalty, AsyncFC early-terminates decoding as soon as the \texttt{await\_future} function name is detected and starts polling for resolved futures. Nevertheless, the runtime still needs to wait for enough tokens to recognize the \texttt{await\_future} function name. This overhead is dominated by TTFT and token generation latency, and is especially pronounced in black-box API models such as GPT-4o, where the runtime cannot inspect internal states to predict the model’s intent in advance.

To further reduce this remaining overhead, AsyncFC supports an optional parallel decoding optimization. 
Rather than waiting for a single decoding stream to reveal whether it will emit \texttt{await\_future}, the runtime may initiate multiple decoding streams from different candidate integration points concurrently and select the first stream that produces the next non-await model action.

\begin{figure}[t]
  \centering
  \begin{subfigure}[t]{0.48\linewidth}
    \vspace{0pt}
    \centering
    \begin{lstlisting}[
        language=Python,
        frame=tb,
        basicstyle=\ttfamily\small,
        breaklines=true,
        columns=fullflexible,
        showstringspaces=false
    ]
@op(
  reads=[
    {"/vehicle/doors": True},
    {"/vehicle/drive": True},
    {"/vehicle/fuel": True}
  ],
  writes=[{"/vehicle/drive": True}],
  session_read=False,
  session_write=False,
  outputs={"engineState": "",
           "fuelLevel": 0.0,
           "batteryVoltage": 0.0}
)
def startEngine(self, ignitionMode: str) \
  -> Dict[str, Union[str, float]]:
    \end{lstlisting}
    \label{fig:code_annotation}
  \end{subfigure}
  \hfill
  \begin{subfigure}[t]{0.42\linewidth}
    \vspace{0pt}
    \centering
    \includegraphics[width=\linewidth]{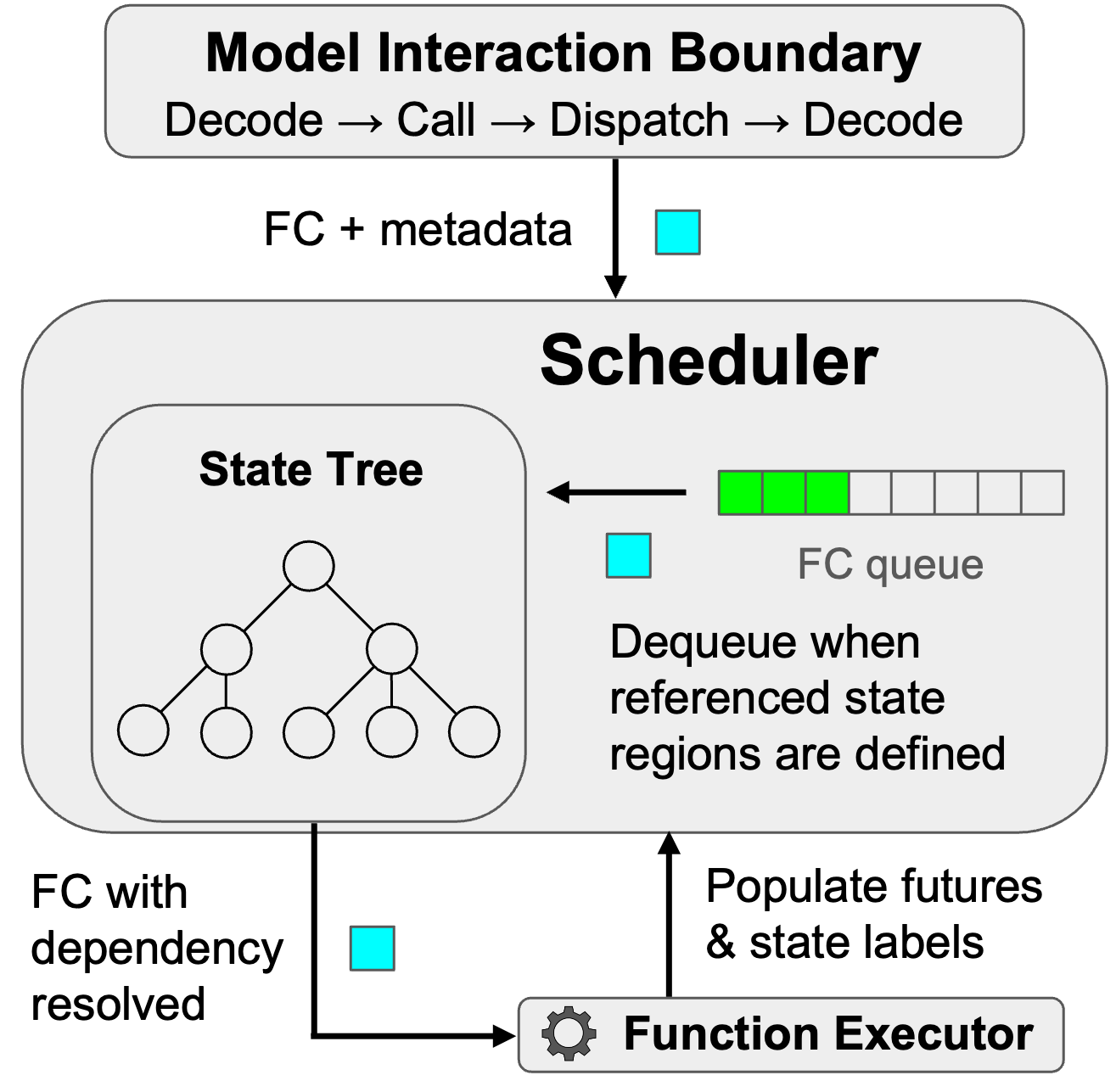}
    \label{fig:scheduling_architecture}
  \end{subfigure}
  \vspace{-2mm}
  \caption{
    \textbf{AsyncFC runtime design.}
    Left: Example of dependency and output structure annotation. Dependency annotations specify the read and write sets, and the runtime automatically infers future ID structure from example values in output schema annotations.
    Right: Overview of the AsyncFC execution pipeline. Model-emitted function calls are synchronously dispatched to the scheduler and enqueued with metadata. The scheduler admits function calls to the State Tree in enqueue order once their referenced state paths are resolved. The State Tree resolves read/write dependencies and dispatches dependency-resolved calls to the executor. Upon completion, function execution populates futures and state labels.
    } 
  \label{fig:neurips_runtime_design}
  
\end{figure}

\subsection{Dependency-Aware Task Scheduling}

Sequential function-calling executes functions in decode order, which implicitly enforces dependencies between calls through blocking. By unblocking model decoding and integrating function execution results at the turn boundary, AsyncFC enables the model to generate subsequent function calls earlier than in the standard blocking execution setting, creating new opportunities for inter-function parallelism.  However, relaxing strict decode-order execution introduces correctness risk, as naive parallelization may execute dependent calls prematurely.

To address this, AsyncFC provides a scheduler that overly conservatively serializes all function executions in decode order by default, enabling only the overlap between model decoding and function execution. To unlock additional inter-function parallelism, the system allows developers to supply optional dependency annotations that explicitly declare function resource accesses. The scheduler uses these annotations to automatically derive function dependencies, permitting concurrent function execution only when safe.

\textbf{Dependency Specification with Labeling.} 
To enable developers to supply these annotations, AsyncFC introduces a lightweight unified labeling mechanism via decorators (Figure~\ref{fig:neurips_runtime_design}). Resources are modeled as a hierarchical state space addressed by path-like identifiers, and function accesses are declared using read/write labels qualified by a scope flag. All annotations are optional; unannotated functions default to root-level read and write sets, 
guaranteeing correctness while still enabling decode–execution overlap. More precise annotations progressively unlock additional inter-function parallelism. Full details, including session-relative and argument-parameterized paths for dependency annotations, are in Appendix~\ref{sec:scheduling_details}.

\textbf{Dependency Enforcement. }
A call depends on prior calls when it accesses overlapping state regions or when its arguments or resource paths use values produced by those prior calls.
To enforce these dependencies, AsyncFC utilizes a runtime scheduler to manage concurrent function execution without modifying the synchronous LLM decoding loop. 
The scheduler dynamically gates function execution through a multi-stage pipeline (Figure~\ref{fig:neurips_runtime_design}). The enforcement mechanism is conceptually related to region-based dependence analysis in Legion \citep{Bauer2012LegionEL} and scoreboarding-style scheduling \citep{thornton1964parallel}.

\textbf{Admission Barrier.} 
Upon decoding, the function call and its metadata are enqueued into the scheduler, while the corresponding future is returned to the LLM. Admission to the State Tree follows the queue order, and a call is admitted only when it reaches the head of the queue and its referenced state paths are resolved. Calls whose paths depend on pending argument or session values remain queued until those values are available.

\textbf{Conflict Analysis.} 
When a function call is admitted, the State Tree records its read and write regions by attaching access labels, which are implemented as unresolved futures. When a new function arrives, the State Tree compares its requested region accesses against these previously registered labels. If access regions overlap and are in conflict, the State Tree attaches the futures of the prior labels as blocking dependencies for the new call. The State Tree dispatches the function to the executor when all its blocking access label futures are resolved. 

\textbf{Execution. }The executor dispatches each admitted call to an independent worker that awaits unresolved argument futures locally before execution. This waiting is non-blocking, and a call awaiting its arguments does not prevent other ready calls from executing. Upon a function's execution completion, the executor fulfills the result future previously returned to the model and removes its access labels from the State Tree, resolving the access label futures and unblocking waiting functions.

\subsection{Error Handling}

When a function call fails, AsyncFC cancels all subsequent function calls that are transitively dependent on the failed operation while keeping function calls that are independent of the failed operation untouched. AsyncFC then uses user messages to inject failure return messages for the failed call and all canceled calls at the turn boundary. The failure messages explicitly distinguish between an operation’s own execution error and cancellation due to dependency on a failed operation. Failure messages integrated after the final decode turn trigger additional recovery turns. 

\subsection{Generalization to Asynchronous Thinking}

Other classes of long-latency LLM operations, such as asynchronous thinking, can be realized as direct specializations of the AsyncFC execution framework.
Conceptually, asynchronous thinking is a paradigm where the main model acts as a coordinator, delegating sub-tasks to independent thinking agents and asynchronously gathering their results to form a final solution.
AsyncFC can enable this paradigm by treating each delegated thinking turn as a long-latency function call, where the argument encodes a subquery along with relevant context, and the return value corresponds to the thinking result. The thinking function issues a model invocation, while the decoding, result integration, and execution of this function follow the same flow as standard functions. This design also allows asynchronous thinking to interleave with asynchronous function execution.

\section{Design Intuition}
\label{sec:Design_Intuition}
The speedup achievable by AsyncFC is governed by the task structure. We represent this structure as a Task-specific Function Dependency DAG, where nodes represent function calls, and edges represent the causal constraints required to complete the workload. These constraints arise from data and state dependencies across calls, or the necessity for the model to observe the execution result of one call before generating the next. 
Let $T_{\text{tool}}$ denote the total function execution time summed across all nodes, and let $T_{\text{cp}}$ denote the sum of function execution times along the critical path. Under ideal asynchronous execution, the wall-clock time spent on function execution is reduced from $T_{\text{tool}}$ to $T_{\text{cp}}$, and this critical-path execution time fully overlaps with total decoding time $T_{\text{LLM}}$. Thus, the ideal asynchronous latency is $\max(T_{\text{LLM}}, T_{\text{cp}})$, giving the theoretical maximum speedup ratio
\begin{equation}
\label{eq:ideal_speedup}    R=\frac{T_{\text{LLM}}+T_{\text{tool}}}
        {\max(T_{\text{LLM}},T_{\text{cp}})} 
\end{equation}
The necessary condition for any speedup is that the model can generate function calls without waiting for all pending executions to complete. Large speedups further
require exploitable function parallelism, i.e. $T_{\text{tool}} \gg T_{\text{cp}}$, and enough overlap between decoding and the critical-path execution duration, ideally $T_{\text{LLM}} \approx T_{\text{cp}}$. Model decode latency also affects
the realized speedup. When tool critical-path time dominates, additional decoding time can be hidden behind execution, whereas when decoding dominates, slower models reduce the attainable speedup. Finally, the total latency savings can be decomposed into the sum of decode-execution overlap and inter-function parallelism. Full derivations, model-latency sensitivity analysis, and the formal savings decomposition are provided in Appendix~\ref{sec:theory}.

\section{Evaluation}
\label{sec:evaluation}
 
This section evaluates AsyncFC along four dimensions: correctness, end-to-end latency, latency scaling, and annotation robustness. We further study its generality on software engineering and asynchronous-thinking workloads.

\begin{figure}[t]
  \centering
  \begin{subfigure}[b]{0.49\linewidth}
    \centering
    \footnotesize
    \textbf{BFCL v3 accuracy.}\par\medskip
    \resizebox{\linewidth}{!}{%
    \begin{tabular}{lccccc}
      \toprule
      \multicolumn{2}{c}{Sync} & \multicolumn{3}{c}{AsyncFC} \\
      \cmidrule(lr){1-2} \cmidrule(lr){3-5}
      Seq. FC & Par. FC & No-Ann(S) & (S) & (P) \\
      \midrule
      68.0\% & 67.3\% & 70.7\% & 69.3\% & 66.0\% \\
      \bottomrule
      \vspace{-1mm}
    \end{tabular}}
    \textbf{BFCL v4 Web Search.}\par\medskip
    \resizebox{\linewidth}{!}{%
    \begin{tabular}{lcccc}
      \toprule
      Metric & Seq. FC & AsyncFC (S) & Par. FC & AsyncFC (P) \\
      \midrule
      Acc. & 54.8\% & 53.2\% & 50.0\% & 50.0\% \\
      Speedup & -- & 1.26$\times$ & -- & 1.12$\times$ \\
      \bottomrule
    \vspace{2mm}
    \end{tabular}}
    \label{fig:neurips_bfcl_tables_panel}
  \end{subfigure}
  \hfill
  \begin{subfigure}[b]{0.50\linewidth}
    \centering
    \includegraphics[width=\linewidth]{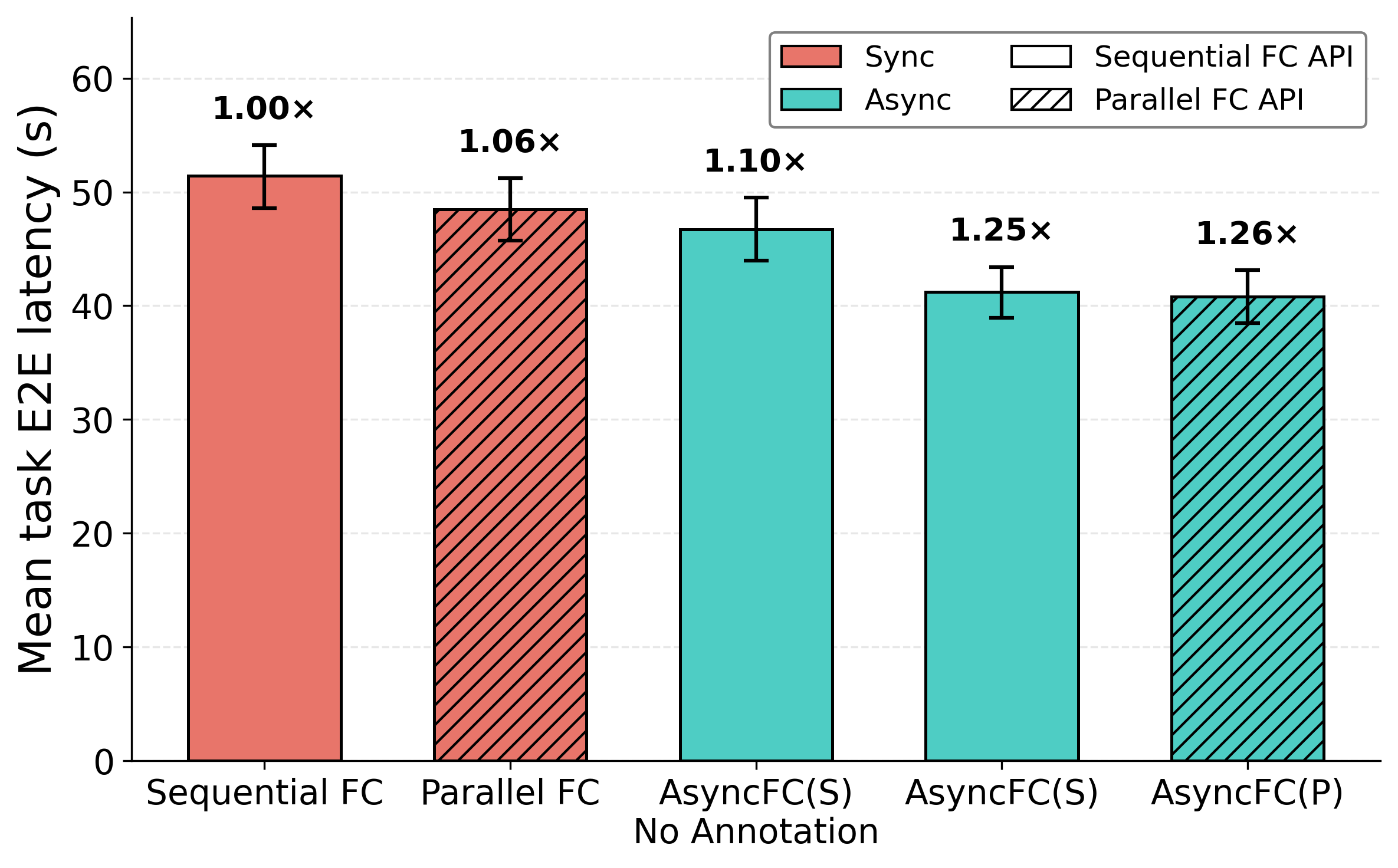}
    \label{fig:neurips_speedup_panel}
    \vspace{-3mm}
  \end{subfigure}
  \caption{\textbf{Main BFCL Results.} Results are reported on BFCL v3 Multi-Turn ($n=150$, 5s delay, GPT-4o) and BFCL v4 Web Search with real backend latency (matched non-overflow composed workloads: $n=31$ for Sequential FC and $n=29$ for Parallel FC, GPT-4o). AsyncFC shows no evidence of a statistically significant accuracy difference ($p_{\mathrm{acc}}>0.05$) and achieves speedups in all settings, with statistically significant latency reductions ($p_{\mathrm{lat}}<0.05$) in all but AsyncFC(P) on BFCL v4 Web Search relative to Parallel FC. 
  Left: Accuracy and BFCL v4 speedup summaries.
  Right: Mean per-task end-to-end latency on BFCL v3, showing AsyncFC(S) outperforms Parallel FC while AsyncFC without annotation still reduces latency. Numbers above bars indicate speedup relative to Sequential FC, and error bars denote 95\% bootstrap confidence intervals over matched cases.}
  \label{fig:neurips_bfcl_summary}
\end{figure}

\subsection{Experimental Setup}
\label{subsec:Experiment_setup}
Across function-calling settings, Sequential FC and Parallel FC are the synchronous baselines using the sequential and parallel function-calling APIs, respectively. In Parallel FC, same-turn function calls are executed concurrently, but the next decoding turn blocks until all results return. AsyncFC(S) and AsyncFC(P) apply AsyncFC on top of the sequential and parallel function-calling APIs, respectively, as pure execution-layer replacements for their corresponding baselines.

For underlying models, BFCL, annotation-robustness, and asynchronous-thinking experiments use GPT-4o. Cross-model experiments use Gemini 3.1 Pro, and SWE-bench Lite experiments use GPT-5.6 Sol in the priority tier. To limit compute costs, we disable optional parallel decoding. Results within each panel are computed from the same run. While latency and accuracy evaluate the efficiency and correctness of AsyncFC, accuracy also provides evidence that LLMs can successfully reason over unresolved future placeholders returned by functions. 

Due to the model providers' ambiguity in function execution order for Parallel FC (Section~\ref{sec:related_work}), we adopt a parallel execution with serial semantics for Parallel FC in figures other than Figure~\ref{fig:neurips_bfcl_summary}. Same-turn functions are executed fully in parallel, but their effects on state mutation are applied in the serial model-emitted order. This assumption is favorable to Parallel FC because it preserves the latency benefit of overlapping same-turn function executions while avoiding accuracy loss from race-dependent state updates that would otherwise produce nondeterministic function returns.

\textbf{Statistical Tests.}
Unless otherwise stated, reported speedups compare each AsyncFC variant against its corresponding synchronous baseline under the same function-calling API; exceptions specify their reference baseline in the figure or table caption.
For latency, we perform a one-sided paired $t$-test on per-task log speedups against the corresponding synchronous baseline and report significance value $p_{\text{lat}}$ for the alternative hypothesis that the geometric mean speedup exceeds 1.
For accuracy, we use McNemar's $\chi^2$ test~\citep{mcnemar1947note}, with significance value $p_{\text{acc}}$, to test for statistically significant differences in paired task outcomes against the corresponding synchronous baseline.

\subsection{BFCL Augmentation and Results}
Speedup is fundamentally limited by workload structure. Many benchmarks use dummy backends with negligible function latency and sequential tasks that lack intrinsic concurrency, underrepresenting AsyncFC's benefits \citep{kulkarni2025massive, qin2023toolllm, ye2025tooleyes, zhuang2023toolqa}. Real-world workloads, however, often involve interleaved tasks and significant function execution delays. To evaluate this, we augment BFCL in two ways. 
For BFCL v3 experiments, we use a filtered evaluation set of 150 multi-turn
cases after removing instances with quality issues. We inject a 5s per-function
delay because the original backend functions have negligible latency. To evaluate performance against a real backend, we utilize the BFCL v4 Web Search split. Because these web search tasks exhibit strong sequential dependencies, we take the 100 raw cases and compose each pair of cases into a single workload to surface inter-task parallelism. Since the composition causes some trajectories to exceed the model's context limit, we report accuracy and latency on the subset of cases that do not overflow in either the synchronous baseline or its AsyncFC counterpart.

Figure~\ref{fig:neurips_bfcl_summary} reports the task-level accuracy and end-to-end latency results. Across all BFCL conditions, AsyncFC preserves accuracy and achieves speedups over both corresponding synchronous baselines in
all settings. Latency gains are
statistically significant in all but one comparison, AsyncFC(P) on BFCL v4 Web Search relative to Parallel FC due to variability in  experiments. With dependency annotations, AsyncFC on BFCL v3 layered over the sequential FC API achieves a larger speedup than native Parallel FC, while
the no-annotation setting still reduces latency by overlapping decoding with serialized function execution. 

\begin{figure}[t]
  \centering
  \begin{subfigure}[t]{0.49\linewidth}
    \vspace{0pt}
    \centering
    \includegraphics[width=\linewidth]{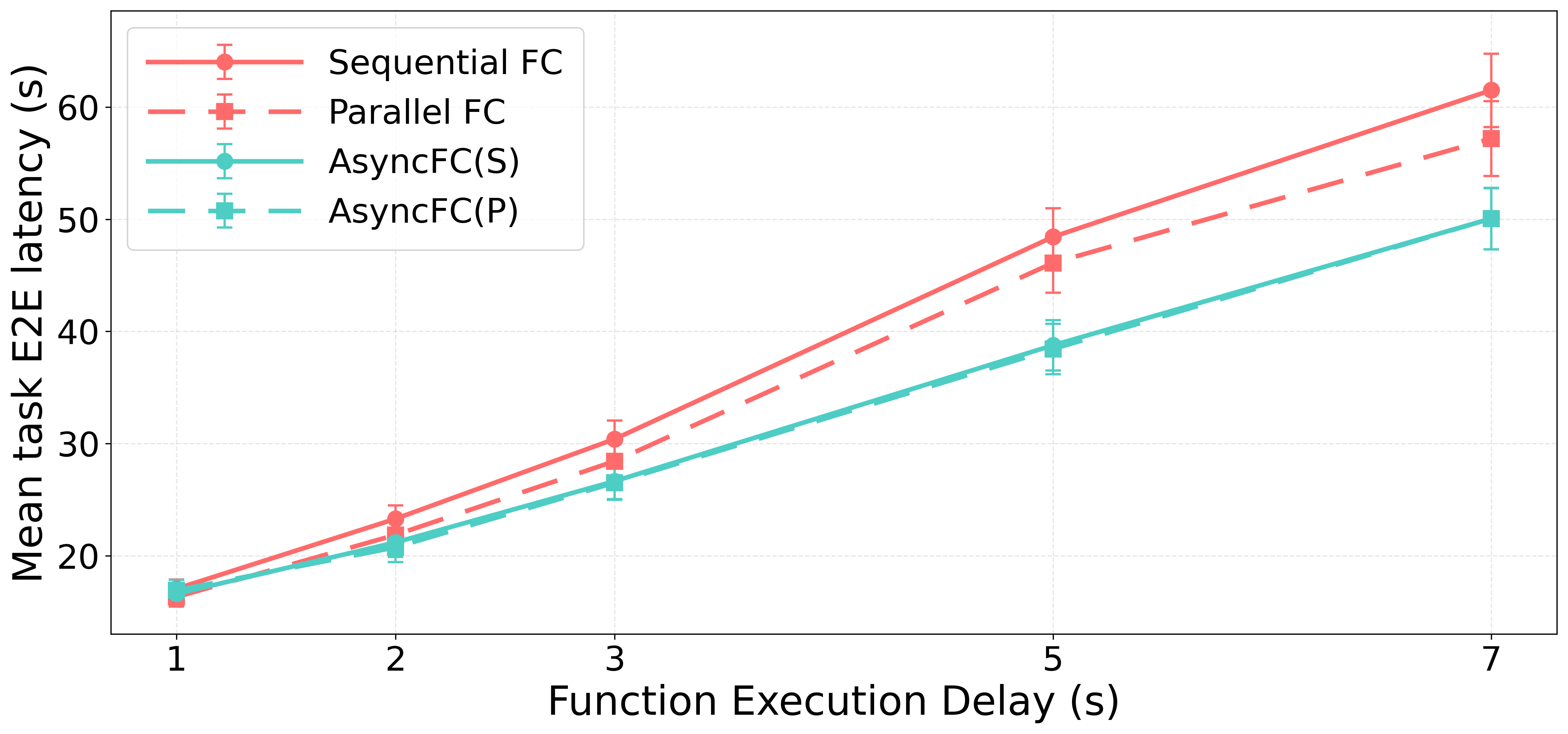}
    
    \label{fig:neurips_latency_sweep_panel}
  \end{subfigure}
  \hfill
  \begin{subfigure}[t]{0.49\linewidth}
    \vspace{0pt}
    \centering
    \includegraphics[width=\linewidth]{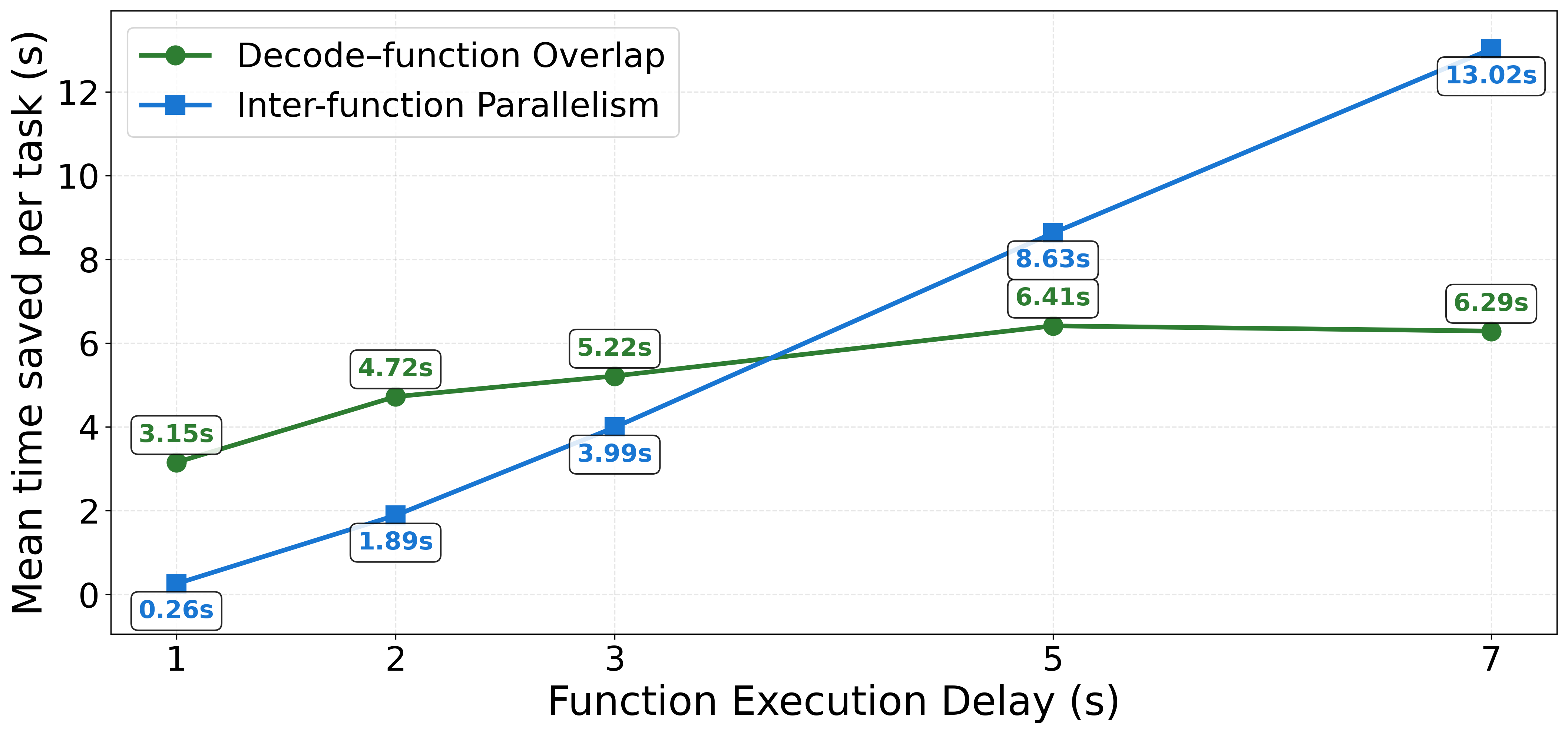}
    
    \label{fig:neurips_two_stage_panel}
  \end{subfigure}
  
  \caption{\textbf{BFCL latency-sweep analysis.}
  Results are reported on BFCL v3 Multi-Turn ($n=150$, GPT-4o) while varying injected per-function delay. The left panel shows mean task end-to-end latency, and the right panel decomposes AsyncFC(S) savings over Sequential FC into decode--execution overlap and inter-function parallelism. Error bars denote 95\% bootstrap confidence intervals obtained by resampling matched cases. AsyncFC shows no evidence of statistically significant accuracy degradation across the sweep (all $p_{\mathrm{acc}} > 0.05$). The decomposition shows that decode--function overlap drives the initial gains, while inter-function parallelism dominates at larger function latencies.}
  \label{fig:neurips_latency_analysis}
  \vspace{-2mm}
\end{figure}

\textbf{Latency-Sweep Analysis.} To isolate where these gains come from and how they scale with function latency, we next perform a controlled latency study on BFCL v3. Following the analysis in Appendix~\ref{subsec:Saving_Decomposition}, we decompose the latency savings into their fundamental components. In net task end-to-end latency, these component savings are partially offset by AsyncFC-specific overhead. Figure~\ref{fig:neurips_latency_analysis} shows a two-stage trend within the evaluated latency range: decode–function execution overlap provides the initial gains and begins to saturate, after which inter-function parallelism becomes the dominant contributor as function latency increases. 

\subsection{Generalization and Robustness}
\label{subsec:Generalization_and_Robustness}

\textbf{Cross-Model Generalization.}
We evaluate AsyncFC using Gemini 3.1 Pro on BFCL v3 Multi-Turn test cases. Because Gemini exhibits a per-turn decoding time approximately $2\times$ slower than GPT-4o, we proportionally scale the simulated tool latency for this evaluation to 10 seconds, as characterized in Equation~\eqref{eq:ideal_speedup}. AsyncFC achieves speedup while maintaining accuracy (Figure~\ref{fig:neurips_additional_evals}).

\textbf{Annotation Robustness.} Because dependency annotations act as safety contracts, any uncertainty about function resource footprints—whether due to complex function logic or a desire to minimize developer effort—is handled through conservative over-estimation. We evaluate AsyncFC robustness without user-provided annotations by falling back to root-level read and write sets, yielding a fully automatic but conservative scheduler that serializes tool execution. As shown in Figure~\ref{fig:neurips_bfcl_summary},  AsyncFC yields speedup by overlapping model decoding with conservative function execution, while preserving baseline accuracy. We also use an external LLM to do one-time offline dependency annotation. The generated
annotations recover additional parallelism over the conservative no-annotation baseline, with annotation-quality metrics and end-to-end results reported in Appendix~\ref{app:llm_annotations}.

\subsection{Downstream Applications}
\textbf{Software Engineering.} We extend our evaluation to SWE-bench Lite by integrating AsyncFC into the SWE-agent. We first transform the underlying function schemas to allow the model to reason about futures and equip the agent with its full original suite of functions. For safe concurrent execution, we implement a rule-based dependency analyzer that uses pattern matching over command names and arguments to identify and provide dependency annotations for functions. If no rule is matched, the system conservatively falls back to locking the root directory. Furthermore, our current rules are intentionally conservative for simplicity and generalization across repositories.
For example, the rule set specifies that \texttt{Python} and \texttt{pytest} lock the root path to prevent conflicts. However, because scripts invoked by these functions usually touch only a minor subset of files, real-world deployments tailored to specific repositories can provide the function with fine-grained operation metadata about file accesses to bypass these coarse root locks and unlock significantly more parallelism. These dynamically labeled functions are then dispatched using the same AsyncFC runtime and scheduler as in our BFCL experiments. 
Because the default SWE-agent function executor is a synchronous shell environment, we replace it with a concurrent subprocess dispatcher that executes each command as an isolated OS subprocess while preserving compatibility with the original command interface.                            

Using GPT-5.6 Sol on SWE-bench Lite, we compare three SWE-agent variants: SWE, the unmodified baseline that enables the Parallel FC API but uses a sequential executor; +Par. FC, which applies the Parallel FC execution semantics defined in Section~\ref{subsec:Experiment_setup}; and +AsyncFC, which integrates AsyncFC. 
AsyncFC achieves a 1.4$\times$ speedup over the SWE-agent baseline and a statistically significant latency improvement over +Par. FC, while preserving issue-resolution rate (Figure~\ref{fig:neurips_additional_evals}).

\textbf{Asynchronous Thinking: Thinking as a Tool. }We further evaluate AsyncFC on asynchronous thinking by treating external reasoning turns as tool invocations. We evaluate this setting on the HotpotQA benchmark, where the model decomposes each task into multiple subqueries and invokes the thinking tool with the subquery and relevant context as arguments. The tool returns the reasoning result, which is then integrated back into the ongoing decoding process. We compose 100
raw tasks into 50 paired workloads. AsyncFC preserves accuracy while providing end-to-end speedups (Figure~\ref{fig:neurips_additional_evals}). These results demonstrate that AsyncFC extends beyond external functions to asynchronous thinking.

\begin{figure}[t]
  \centering
  \begin{subfigure}[t]{0.31\linewidth}
    \vspace{0pt}
    \centering
    \scriptsize
    \textbf{Gemini 3.1 Pro on BFCL v3 Multi-Turn}\par\medskip
    \setlength{\tabcolsep}{3pt}
    \resizebox{\linewidth}{!}{%
    \begin{tabular}{lcc}
      \toprule
      Metric & Par. FC & AsyncFC(P) \\
      \midrule
      Acc. & 62.0\% & 65.3\% \\
      Speedup & -- & 1.17$\times$ \\
      \bottomrule
    \end{tabular}}
    \label{fig:neurips_cross_model_panel}
  \end{subfigure}
  \hfill
  \begin{subfigure}[t]{0.33\linewidth}
    \vspace{0pt}
    \centering
    \scriptsize
    \textbf{SWE-Bench Lite}\par\medskip
    \setlength{\tabcolsep}{3pt}
    \resizebox{\linewidth}{!}{%
    \begin{tabular}{lccc}
      \toprule
      Metric & SWE & +Par. FC & +AsyncFC \\
      \midrule
      Acc. & 79.0\% & 79.0\% & 79.3\% \\
      \makecell[l]{Speedup\\vs. SWE} & -- & 1.04$\times$ & 1.40$\times$ \\
      \bottomrule
    \end{tabular}}
    \label{fig:neurips_swe_panel}
  \end{subfigure}
  \hfill
  \begin{subfigure}[t]{0.31\linewidth}
    \vspace{0pt}
    \centering
    \scriptsize
    \textbf{HotpotQA}\par\medskip
    \setlength{\tabcolsep}{3pt}
    \resizebox{\linewidth}{!}{%
    \begin{tabular}{lcc}
      \toprule
      Metric & Seq. FC & AsyncFC(S) \\
      \midrule
      Acc. & 75.0\% & 75.0\% \\
      Speedup & -- & 1.24$\times$ \\
      \bottomrule
    \end{tabular}}
    \label{fig:neurips_thinking_panel}
  \end{subfigure}
  \caption{
  \textbf{Cross-model and downstream application evaluations.} The three panels report cross-model transfer ($n=150$, Gemini 3.1 Pro, 10s delay), software-engineering results ($n=300$, GPT-5.6 Sol, real function latency), and asynchronous-thinking results (composed workloads, $n=50$, GPT-4o, real function latency), respectively. Speedups are relative to the corresponding synchronous baseline. For SWE-Bench Lite, displayed speedups are relative to SWE-agent for a common scale; latency significance tests compare +AsyncFC against +Par. FC. Across all settings, AsyncFC achieves statistically significant speedups, without evidence of statistically significant accuracy degradation (all $p_{\mathrm{lat}} < 0.05$, all $p_{\mathrm{acc}} > 0.05$).}
  \label{fig:neurips_additional_evals}
  \vspace{-5mm}
\end{figure}

\section{Discussion and Limitations}
\label{Sec:Discussion_Limitation}
\textbf{Model Capability Discovery.}
AsyncFC requires the model to tolerate unresolved symbolic references. Across evaluations, AsyncFC preserves accuracy, providing evidence that LLMs can natively reason over unresolved future placeholders in our future-based context. We hypothesize that this behavior arises from model priors learned from large corpora of software artifacts, including asynchronous patterns (e.g., futures/promises and non-blocking I/O) that routinely involve symbolic handles that are resolved later.
Such native support for asynchronous model–tool interaction unlocks a new paradigm for designing efficient and flexible agentic workflows. 

\textbf{Workload Dependence.} As characterized in Section~\ref{sec:Design_Intuition} and Appendix~\ref{sec:theory}, AsyncFC's speedup depends on the workload structure. Strictly sequential tasks and function calls with negligible latency limit AsyncFC's speedup. Additionally, AsyncFC can incur decoding overhead, such as when a function completes during the \texttt{await\_future} decoding. 

\textbf{Application Scenarios}. AsyncFC is ideal for long-latency functions where immediate execution outputs aren't needed for decode progress, such as "write-only" actions (e.g., booking tickets, sending emails). It also may benefit cyber-physical systems and robotics by overlapping reasoning with physical execution to amortize real-world delays.

\textbf{Potential Optimization.} While sequential models generate function calls in a fixed linear sequence, this output represents just one valid linearization of the underlying Task-specific Function Dependency DAG. Consequently, multiple decoding orders may be consistent with the task's logical constraints. A potential optimization is to employ beam search to explore these alternative trajectories in parallel. By selecting decoding paths that immediately produce executable functions—those whose dependencies are already satisfied—the system optimizes the concurrent execution throughput.

\textbf{Systems Side Effects.}
AsyncFC can improve KV-cache utilization by reducing the duration for which request-specific KV state remains in GPU memory. 
By overlapping decoding with function execution, AsyncFC shortens the wall-clock lifetime of each request, thereby reducing the mean residency time of its KV cache.

\section{Conclusion}
As LLM agents increasingly operate in environments with rich internal parallelism and non-negligible function latency, strict synchronous execution becomes a key bottleneck. 
We present AsyncFC, a future-based asynchronous function-calling execution-layer framework that decouples model decoding from function execution while preserving the standard function-calling protocol. By leveraging futures and dependency-aware scheduling, AsyncFC safely enables decoding-execution overlap and exposes inter-function parallelism when dependencies permit. 
Without modifying the model, protocol, or synchronous function implementation, AsyncFC reduces end-to-end task latency and preserves accuracy. Beyond these efficiency gains, our results provide evidence that modern LLMs can natively reason over future-valued placeholders. This suggests an asynchronous model–tool interaction pattern that can be leveraged for more flexible and efficient agentic workflows.

\begin{ack}
We would like to thank the anonymous reviewers for their constructive feedback, as well as the members of the Sky Computing Lab and Lab11 for their valuable discussions. We also appreciate support from James Lim for assistance in adapting AsyncFC into SWE-agent, and the assistance of Yijia Wu (University of Rochester) and Kexin Liu in expanding the experimental evaluations. Sky Computing Lab is supported by gifts from Accenture, Amazon, AMD, Anyscale, Broadcom, cmpnd, Google, IBM, Intel, Intesa Sanpaolo, Lambda, Lightspeed, Mibura, NVIDIA, Samsung SDS, and SAP.
\end{ack}

{
\small



\bibliographystyle{abbrvnat}
\bibliography{references}
}


\appendix

\section{Details for Dependency Specification with Labeling.} 
\label{sec:scheduling_details}
To enable developers to supply these annotations, AsyncFC introduces a lightweight unified labeling mechanism via decorators (Figure~\ref{fig:neurips_runtime_design}). While this decorator also manages output schemas (Section~\ref{subsec:Nonblocking_Decoding}), its primary role in scheduling is to provide dependency annotations.

\textbf{Hierarchical Resource Paths.} AsyncFC models resources as a logical, possibly hierarchical state space, structured and addressed using path-like resource identifiers. Function calls are interpreted as reading from or writing to named regions within this state space, identified by these resource paths. Accordingly, read and write labels in dependency annotation explicitly represent each function call’s access intent by enumerating the regions of state that the call may observe or mutate. Each read or write label is further qualified by a scope flag that specifies whether the access applies only to the targeted resource path or conservatively extends to all descendant paths in the state space.

\textbf{Robustness to Partial Annotation.}
All annotations are optional and may be provided independently. For unannotated functions, the runtime defaults to root-level read and write sets, serializing function execution while still allowing model decoding to overlap with function execution. 
This conservative strategy guarantees out-of-the-box correctness without requiring any annotations. To move beyond this baseline, AsyncFC enables an incremental path to performance. Partial annotations unlock some inter-function parallelism, while more precise annotations progressively expose additional opportunities for concurrent execution and further speedups.

\textbf{Session-Relative Paths.} While many resource accesses can be expressed using absolute paths in the state space, some function calls operate on resources whose exact locations are dependent on prior operations. To support such cases, AsyncFC allows dependency labels to be expressed using session-relative paths. A session label binds a function call’s access to the current session state—implicitly derived from prior function executions—enabling paths to be resolved dynamically based on the effects of earlier operations. To label session access, a function call that reads session state only declares its dependence on the current session context, whereas a function call that writes to session state additionally specifies how the session advances.
This advancement induces an ordering constraint on subsequent functions that access the session.

\textbf{Argument-Parameterized Paths. }
In addition to session-relative addressing, some function calls operate on resources whose exact locations are determined by the arguments of the current invocation. AsyncFC therefore allows dependency labels to be parameterized by function arguments, enabling read and write paths to be derived after the call is decoded and submitted to the runtime. 

\section{Design Intuition: Operation Regimes and Speedup Patterns}
\label{sec:theory}
This section expands the speedup analysis summarized in Section~\ref{sec:Design_Intuition}. We model each task as a Task-specific Function Dependency DAG, where nodes represent function calls and edges represent the causal constraints required to complete the workload. Let $T_{\text{tool}}$ denote the total function execution time summed across all nodes, and let $T_{\text{cp}}$ denote the sum of function execution times along the critical path.

\textbf{Maximum Speedup Bound.} We restate the maximum speedup bound in Equation~\eqref{eq:ideal_speedup} to support the appendix analysis.
Under ideal asynchronous execution, the wall-clock time spent on function execution is reduced from $T_{\text{tool}}$ to $T_{\text{cp}}$, and this critical-path execution time fully overlaps with total decoding time $T_{\text{LLM}}$. Consequently, the
end-to-end latency is
$\max\!\left(T_{\text{LLM}},\,T_{\text{cp}}\right)$. By taking the ratio of the baseline synchronous latency to the asynchronous latency, the theoretical maximum speedup ratio $R$ is:
\begin{equation}
\label{eq:appendix_ideal_speedup}    R=\frac{T_{\text{LLM}}+T_{\text{tool}}}
        {\max(T_{\text{LLM}},T_{\text{cp}})} 
\end{equation}

\begin{figure}[t]
  \centering

  \captionof{table}{\textbf{AsyncFC operation regimes.} These regimes are induced by inter-function parallelism and decode-function overlap,
  under the necessary condition that decode progress does not depend on all executing functions' results.} 
  \label{tab:asyncfc-regimes}
    \begin{center}
    \begin{small}
      \begin{sc}
        \begin{tabular}{lcc}
          \toprule
          \diagbox{$T_{\text{LLM}}/T_{\text{cp}}$}{$T_{\text{tool}}/T_{\text{cp}}$}
          & $ \approx 1$ 
          & $ \gg 1$ \\
          \midrule
          $ \ll 1 \ \text{or} \ \gg 1$
          & Little 
          & Moderate \\
          $ \approx 1$
          & Moderate 
          & \textbf{sweet spot} \\
          \bottomrule
        \end{tabular}
      \end{sc}
    \end{small}
  \end{center}

  \vspace{1mm}


    \includegraphics[width=\columnwidth]{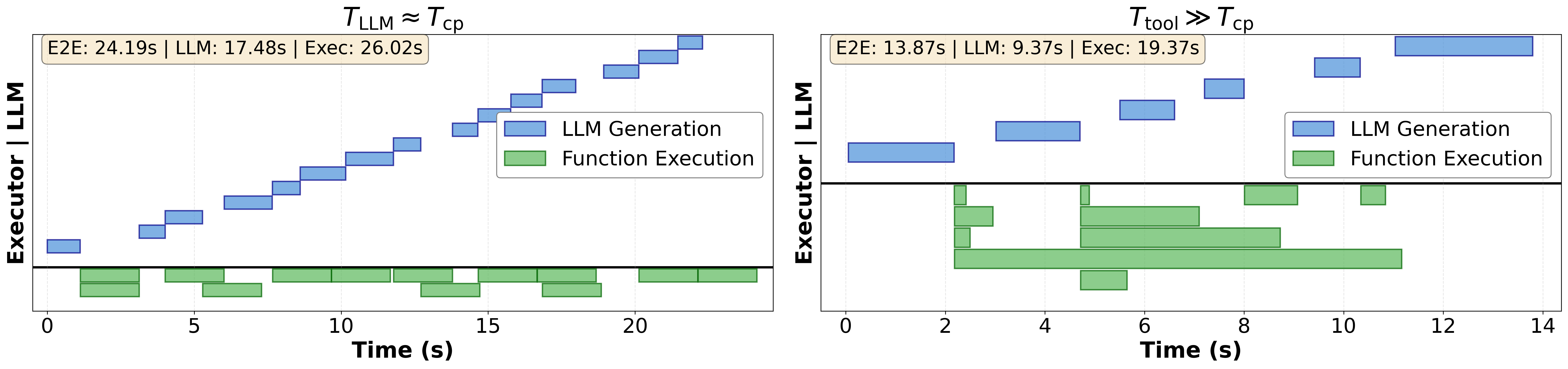}
    \captionof{figure}{
    \textbf{Real workload execution traces under distinct AsyncFC operation regimes.}
    Left: Balanced model decode and function critical-path times ($T_{\text{LLM}} \approx T_{\text{cp}}$), where decoding with futures enables continuous decoding.
    Right: High inter-function parallelism ($T_{\text{tool}} \gg T_{\text{cp}}$), where short-latency function calls allow decoding to proceed.
    }
    \label{fig:asyncfc-traces}
  \vspace{-3mm}
\end{figure}

\textbf{Conditions for Attaining Saving.}
A necessary condition for any speedup is that the model can generate function calls without waiting for all pending executions to complete. When this condition holds, the magnitude of the attainable speedup further depends on two factors. First, there must exist substantial parallelism in the Task-specific Function Dependency DAG (i.e. $T_{\text{tool}} \gg T_{\text{cp}}$), which in practice also requires that the interval between issuing independent function calls remain small relative to execution latencies, so that independent function calls can overlap in time. Second, there is sufficient decoding work and critical-path execution duration to enable substantial overlap (i.e. $T_{\text{LLM}} \approx T_{\text{cp}}$). The operation regimes are summarized in Table~\ref{tab:asyncfc-regimes}.

Figure~\ref{fig:asyncfc-traces} shows two workload traces illustrating the operation regimes and speedup patterns of AsyncFC. In the first trace, the function critical-path time and decoding time are comparable, and decoding with futures satisfies the necessary condition. In the second trace, the Task-specific Function Dependency DAG exhibits substantial parallelism, and decoding the next turn using short function returns while long function calls remain in flight satisfies the necessary condition.

\subsection{Sensitivity to Model Capability}
While Equation \eqref{eq:appendix_ideal_speedup} defines the theoretical speedup ceiling, actual asynchronous execution introduces decoding overhead, modeled as a fractional increase $\alpha$ in LLM decoding time. With this overhead, the speedup ratio becomes:
\[
    R = \begin{cases} 
      \displaystyle \frac{T_{\text{LLM}} + T_{\text{tool}}}{T_{\text{cp}}}, & \text{if } T_{\text{cp}} \ge (1 + \alpha) T_{\text{LLM}} \\[1em]
      \displaystyle \frac{T_{\text{LLM}} + T_{\text{tool}}}{(1 + \alpha) T_{\text{LLM}}}, & \text{if } (1 + \alpha) T_{\text{LLM}} > T_{\text{cp}}
   \end{cases}
\]
This expression reveals two workload-dependent regimes. When the Task-specific Function Dependency DAG is deep or contains long-latency operations such that the critical path exceeds the model's total decoding time, a larger model with greater total decoding time $T_{\text{LLM}}$ improves the speedup. By contrast, for a workload with a shallow DAG (small $T_{\text{cp}}$), a larger model leads to a smaller speedup. Beyond decoding overhead, asynchronous function calling may occasionally introduce additional end-to-end overhead by decoding more turns than a synchronous baseline.

\subsection{Saving Decomposition}
\label{subsec:Saving_Decomposition}
Having characterized when large speedups are theoretically attainable, we next analyze how these gains are realized in practice. 
Consider an agent trace for a single task. Let $\mathcal{M}$ and $\mathcal{E}$ denote the sequence of time intervals for model decoding and function execution, respectively. Any such sequence of time intervals can be represented as $\mathcal{I} = \{[s_i, e_i]\}$, where $s_i$ and $e_i$ denote the wall-clock start and end times of the $i$th interval. We define the \textbf{Sequential sum} $\mathcal{S}(\mathcal{I}) = \sum_i (e_i - s_i)$ as the cumulative duration of the intervals, and the \textbf{effective union} $\mathcal{D}(\mathcal{I}) = \left| \bigcup_i [s_i, e_i] \right|$ as the total wall-clock duration covered by the merged intervals. Because model decoding is sequential within a single agent trace, intervals in $\mathcal{M}$ do not overlap and $\mathcal{D}(\mathcal{M})=\mathcal{S}(\mathcal{M})$. 
In this trace, there are two distinct sources of savings. First, \textbf{inter-function parallelism}, $\Delta_{F \parallel F} = \mathcal{S}(\mathcal{E}) - \mathcal{D}(\mathcal{E})$, captures the reduction in execution time due to parallel function execution. Second, \textbf{decode-execute parallelism}, $\Delta_{D \parallel E} = \mathcal{D}(\mathcal{M}) + \mathcal{D}(\mathcal{E}) - \mathcal{D}(\mathcal{M} \cup \mathcal{E})$, captures the end-to-end time reduction gained by overlapping model decoding with function execution. The total latency saving, $T_{\text{saving}}$, defined as the difference between the serialized baseline and the observed end-to-end asynchronous latency, is exactly equal to the sum of these two components:
\[
    T_{\text{saving}} \triangleq \mathcal{S}(\mathcal{M}) + \mathcal{S}(\mathcal{E}) - \mathcal{D}(\mathcal{M} \cup \mathcal{E}) = \Delta_{F \parallel F} + \Delta_{D \parallel E}
\]

\begin{figure}[t]
  \centering
  \begin{minipage}[c]{0.48\linewidth}
    \centering
    \small
    \setlength{\tabcolsep}{5pt}
    \begin{tabular}{lc}
      \toprule
      Metric & Value \\
      \midrule
      Accuracy                        & 0.925 \\
      Recall                          & 0.791 \\
      Precision                       & 0.791 \\
      Over-serialization (FP / total) & 0.038 \\
      Missed dependency (FN / total)  & 0.038 \\
      \bottomrule
    \end{tabular}
    \par\vspace{6pt}    \label{fig:neurips_annotation_accuracy_panel}
  \end{minipage}
  \hfill
  \begin{minipage}[c]{0.48\linewidth}
    \centering
    \small
    \setlength{\tabcolsep}{5pt}
    \begin{tabular}{lcc}
      \toprule
      Metric   & Par.\ FC & \shortstack{AsyncFC\\(LLM Ann.)} \\
      \midrule
      Accuracy & 68.0\%   & 68.0\%       \\
      Speedup  & --       & 1.22$\times$ \\
      \bottomrule
    \end{tabular}
    \par\vspace{6pt}
    \label{fig:neurips_llm_e2e_panel}
  \end{minipage}
  \captionof{table}{\textbf{LLM-generated dependency annotation results.} The left table reports annotation quality on 1812 BFCL v3 Multi-Turn intra-class function pairs by comparing LLM-generated annotation against human-annotated ground truth. The right table reports end-to-end results with GPT-4o on BFCL v3 Multi-Turn under Parallel FC ($n=150$, 5s delay). AsyncFC with LLM-generated annotations achieves statistically significant latency reductions relative to Parallel FC, with no evidence of statistically significant accuracy difference ($p_{\text{lat}} < 0.05$, $p_{\text{acc}} > 0.05$), and its speedup is larger in magnitude than the no-annotation setting in Figure~\ref{fig:neurips_bfcl_summary}.}
  \label{fig:neurips_annotation_results}
\end{figure}

\section{Automatic Annotation Generation.}
\label{app:llm_annotations}
To explore the potential for recovering concurrency beyond conservative fallback without manual annotation effort, in this experiment we use an external LLM (Claude 4.6 Opus) to perform one-time, offline annotation prior to deployment. To evaluate dependency annotation correctness, we extract 1812 distinct, directed function pairs from BFCL Multi-Turn V3. Evaluation is restricted to function pairs within the same class where resource conflicts are theoretically possible, excluding trivial cross-class pairs. For each function pair, we compare the conflict status derived from LLM-generated annotations with that derived from human-annotated ground truth (Table~\ref{fig:neurips_annotation_results}, left). We also apply these annotations to end-to-end evaluation. Right panel of Table~\ref{fig:neurips_annotation_results} shows that they recover additional parallelism beyond the conservative fallback, achieving a speedup larger in magnitude than the no-annotation setting reported in Figure~\ref{fig:neurips_bfcl_summary}.

%

\end{document}